\documentclass[10pt,twocolumn,letterpaper]{article}

\usepackage[pagenumbers]{cvpr} 

\usepackage{graphicx}
\usepackage{amsmath}
\usepackage{amssymb}
\usepackage{booktabs}
\usepackage{multirow}
\usepackage{bbding}
\usepackage[pagebackref,breaklinks,colorlinks]{hyperref}

\usepackage[capitalize]{cleveref}
\crefname{section}{Sec.}{Secs.}
\Crefname{section}{Section}{Sections}
\Crefname{table}{Table}{Tables}
\crefname{table}{Tab.}{Tabs.}
\Crefname{figure}{Figure}{Figures}
\crefname{figure}{Fig.}{Figs.}

\def\confName{CVPR}
\def\confYear{2022}

\begin{document}

\title{Neural MoCon: Neural Motion Control for Physically Plausible Human Motion Capture}

\author{Buzhen Huang\hspace{5mm} Liang Pan\hspace{5mm} Yuan Yang\hspace{5mm} Jingyi Ju\hspace{5mm} Yangang Wang\footnotemark[1]\\%
\\
Southeast University, China
}
\maketitle

\begin{abstract}
   Due to the visual ambiguity, purely kinematic formulations on monocular human motion capture are often physically incorrect, biomechanically implausible, and can not reconstruct accurate interactions. In this work, we focus on exploiting the high-precision and non-differentiable physics simulator to incorporate dynamical constraints in motion capture. Our key-idea is to use real physical supervisions to train a target pose distribution prior for sampling-based motion control to capture physically plausible human motion. To obtain accurate reference motion with terrain interactions for the sampling, we first introduce an interaction constraint based on SDF~(Signed Distance Field) to enforce appropriate ground contact modeling. We then design a novel two-branch decoder to avoid stochastic error from pseudo ground-truth and train a distribution prior with the non-differentiable physics simulator. Finally, we regress the sampling distribution from the current state of the physical character with the trained prior and sample satisfied target poses to track the estimated reference motion. Qualitative and quantitative results show that we can obtain physically plausible human motion with complex terrain interactions, human shape variations, and diverse behaviors. More information can be found at~\url{https://www.yangangwang.com/papers/HBZ-NM-2022-03.html}

\end{abstract}

\renewcommand{\thefootnote}{\fnsymbol{footnote}}
\footnotetext[1]{Corresponding author. E-mail: yangangwang@seu.edu.cn. This work was supported in part by the National Key R\&D Program of China under Grant 2018YFB1403900, the National Natural Science Foundation of China (No. 62076061), the ``Young Elite Scientists Sponsorship Program by CAST" (No. YES20200025), and the ``Zhishan Young Scholar" Program of Southeast University (No. 2242021R41083).}


\section{Introduction}
Recent years have witnessed significant development of marker-less motion capture, which promotes a wide variety of applications ranging from character animation to human-computer interaction, personal well-being, and human behavior understanding. Extensive existing works can kinematically capture accurate human pose from monocular videos and images via network regression~\cite{kanazawa2018end,kolotouros2019learning,kocabas2020vibe,zheng2019deephuman,zhang2020object} or optimization~\cite{pavlakos2019expressive,wang2017outdoor,rempe2021humor,fan2021revitalizing}. However, they are often hard to leverage in real-world systems due to a series of artifacts that are not satisfied biomechanical and physical plausibility~(\eg, jitter and floor penetration). 

To improve motion quality and physical plausibility, a few works focus on capturing human motion using physics-based constraints. \cite{wei2010videomocap,shimada2020physcap,shimada2021neural,xie2021physics,rempe2020contact} incorporate physical laws as soft constraints in numerical optimization framework and reduce artifacts. To make optimization be tractable, they can only adopt simple and differentiable physical models, which may result in high approximation errors. Other methods~\cite{Yu:2021:MovingCam,yuan2021simpoe,peng2018sfv} utilize non-differentiable physics simulators with deep reinforcement learning~(DRL) to achieve accurate and physically plausible 3D human pose estimation. However, training a desirable policy requires complex configurations~\cite{dulac2019challenges,li2017deep,andrychowicz2020matters}, and it may be sensitive to environmental changes~\cite{peng2018deepmimic,Yu:2021:MovingCam}. The limitations above make them be infeasible to estimate human pose with scene interactions and subject varieties for motion capture tasks. Nevertheless, motion control, typically sampling-based methods~\cite{liu2010sampling}, have achieved an impressive performance in reproducing highly dynamic and acrobatic motions and is robust to contact-rich scenarios, which shows a way for general physics-based motion capture.

\begin{figure}
    \begin{center}
    \includegraphics[width=1.0\linewidth]{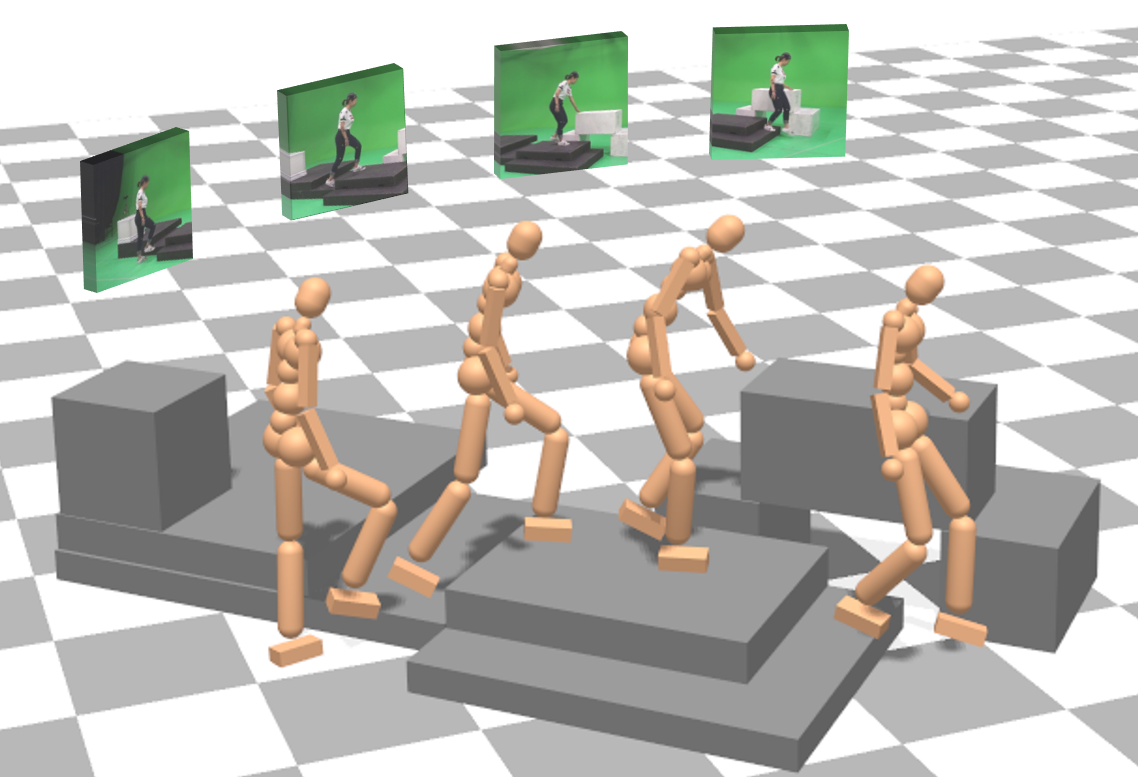}
    \end{center}
    \vspace{-5mm}
    \caption{Our method captures physically plausible human motion from monocular RGB videos via neural motion control.}
    \label{fig:teaser}
    \vspace{-7mm}
    \end{figure}

In this paper, we aim to construct a physics-based motion capture framework that is more general to complex terrains, shape variations, and diverse behaviors along sampling-based motion control. However, employing sampling-based motion control in monocular motion capture tasks faces several challenges. First, conventional sampling-based methods~\cite{liu2015improving,liu2010sampling} often track the accurate reference motion from commercial motion capture systems, while the estimated motion from monocular RGB videos is noisy and physically implausible. An inaccurate contact results in an unnatural pose would even lead to an imbalance state for the character. Second, it is complicated to find an optimal distribution for the sampling. Although CMA~(Covariance Matrix Adaptation)~\cite{hansen2006cma} is proved to be able to adjust distribution with black-box optimization~\cite{liu2015improving}, it requires evaluating plenty of samples for the distribution adaption, which is time-consuming. Furthermore, the adaption relied on random samples from an initial distribution imposes uncertainty for the motion capture.

To address the obstacles, \textbf{our key-idea is to train a motion distribution prior with physical supervisions. The prior provides feasible solutions for sampling-based motion control to capture physically plausible human motion from a monocular color video, which is named as Neural Motion Control (Neural MoCon).} We first introduce a human-scene interaction constraint to obtain a reference motion with appropriate contacts for sampling. 
Different from existing works~\cite{shimada2020physcap,rempe2020contact} to detect foot-ground contact status, our proposed interaction constraint adjusts the distance between two disconnected meshes via SDF, enforcing the human model to be close to the ground surface.

Then, we have tried to train an encoder to regress the distribution with KL divergence~(Kullback-Leibler divergence) and pseudo ground-truth from CMA. However, for the same character state and reference pose, the CMA method obtains different distributions, thus the stochastic error of CMA results in network divergence and erroneous regression. Consequently, we propose a novel two-branch decoder to address this obstacle. As shown in~\cref{fig:distribution}, the target pose sampled from the estimated distribution is fed into a physical branch to verify the validity. Since the simulator is non-differentiable, we use the output to supervise the pose decoder and enforce it to transfer the target pose to a dynamical pose like the simulator. Moreover, a reconstruction loss from the reference pose is applied to the decoded pose to promote correct distribution encoding. When the encoder is convergent, we use it to encode distribution and sample target poses for the physical branch to capture physically plausible motion. The main contributions of this work are summarized as follows.

\begin{itemize}
    \vspace{-2mm}
    \item We propose an explicit physics-based motion capture framework that is more general to complex terrain, body shape variations, and diverse behaviors.
    \vspace{-2mm}
    \item We propose a novel two-branch decoder to avoid stochastic error from pseudo ground-truth and train the distribution prior with a non-differentiable physics simulator. 
    \vspace{-2mm}
    \item We propose an interaction constraint based on SDF to capture accurate human-scene contact from complex terrain scenarios. 
\end{itemize}

\begin{figure*}
    \begin{center}
    \includegraphics[width=1.0\linewidth]{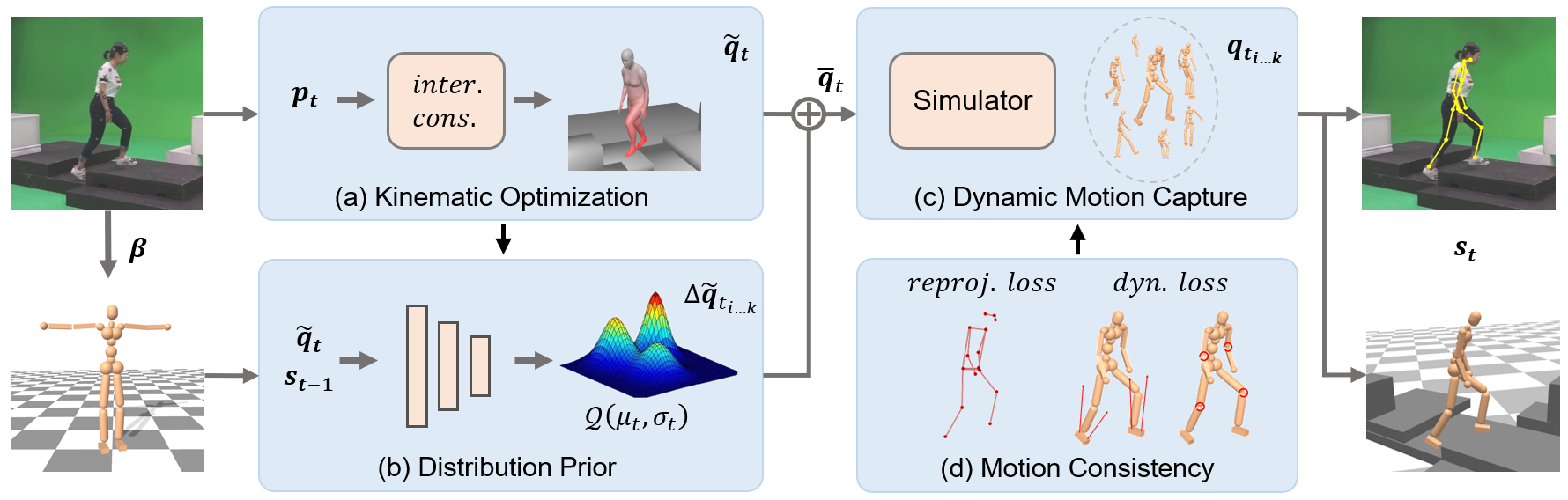}
    \end{center}
    \vspace{-7mm}
    \caption{\textbf{Overview.} Our method first estimates reference motion with accurate human-scene interaction as well as human shape from a monocular RGB video~(a). Then, a prior regresses a distribution from the state of physical character and the reference pose to sample target poses~(b). The physics simulator is used to obtain a physically plausible pose for each sample~(c). The sample with the lowest loss is adopted and used for the next frame after sample evaluation~(d).}
    \label{fig:pipeline}
    \vspace{-7mm}
\end{figure*}

\section{Related Work}\label{sec:relatedwork}
\noindent\textbf{Physics-based motion capture}. VideoMocap~\cite{wei2010videomocap} first employs physical constraints in motion capture by jointly optimizing the human pose and contact force, and this approach requires manual intervention to achieve satisfying results. Based on~\cite{wei2010videomocap}, \cite{li2019estimating} and~\cite{shimada2020physcap,rempe2020contact,zell2017joint} further consider the object interaction and kinematic pose estimation, respectively. Recently, Shimada~\etal~\cite{shimada2021neural} proposed a neural network-based approach to estimate the ground reaction force and joint force and updated the character’s pose using the derived accelerations. To make optimization tractable, their methods can only adopt simple and differentiable physics models with limited constraints, which results in high approximation errors. To address this problem, some latest works~\cite{peng2018sfv, yuan2021simpoe, yuan2020residual, Yu:2021:MovingCam} employ DRL to implement motion capture based on non-differentiable simulators. Nevertheless, training a desirable policy requires complex configurations~\cite{dulac2019challenges,li2017deep,andrychowicz2020matters}, and it may be sensitive to motion types and body shape variations~\cite{peng2018deepmimic,Yu:2021:MovingCam}. Vondrak~\etal~\cite{vondrak2012video} directly used the silhouette to construct a character-image consistency to train a state-machine controller. However, this approach could only be generalized to a variety of motions, and the recovered motion seems to be unnatural. In this work, we adopt neural motion control to capture motion rather than DRL. With the trained distribution prior, our method is more general to different terrain interactions, human shape variations, and diverse behaviors.

\noindent\textbf{Physics-based character control}. Physics-based character control is a longstanding problem~\cite{van1993sensor,wrotek2006dynamo,sharon2005synthesis,lee2014locomotion,lee2021learning,xiang2010physics}. Early works rely on the inverted pendulum model~\cite{kajita1991study}, passive dynamics walking~\cite{kuo2001simple} and zero-moment-point-based trajectory generation~\cite{harada2006analytical} can handle simple motions. To solve large-DOF~(degree-of-freedom) models, optimization-based methods~\cite{kim2008dynamic,xiang2009optimization,sok2007simulating,levine2012physically} are widely used to simulate and analyze human motions. However, it requires substantial computational effort to deal with a complex motion. Other methods~\cite{ye2010optimal,coros2010generalized} approximate the actual human control systems and can produce both normal and pathological walking motions. These control-based methods can generalize to a variety of skills~\cite{coros2010generalized,ye2010optimal,liu2010sampling,liu2015improving,liu2012terrain}, but a set of hyperparameters are required to tune for the desired behaviors. Recent works adopt DRL to control physical character~\cite{peng2018deepmimic,xie2020allsteps,lee2021learning}. It shows that DRL can achieve high-quality motion when motion capture data are provided as a reference~\cite{peng2018deepmimic}. Curriculum learning promotes the DRL to learn more complex tasks~\cite{xie2020allsteps}. However, training an optimal policy takes numerous low- and high-level design decisions, which strongly affect the performance of the resulting agents. We follow sampling-based motion control~\cite{liu2010sampling,liu2015improving} to construct a general framework. Furthermore, we propose a network-based distribution prior to avoid the time-consuming distribution adaption and to improve the stability for their methods.

\noindent\textbf{3D human with scene interaction}. Modeling 3D human with scene interactions will promote the computational understanding of human behavior, which is important for metaverse and related applications. Previous works in scene labeling~\cite{jiang2013hallucinated}, scene synthesis~\cite{fisher2015activity}, affordance learning~\cite{grabner2011makes,kim2014shape2pose} and object arrangement~\cite{jiang2012learning} verified human context is helpful for scene understanding. The prior knowledge of scene geometry can also promote a more reasonable and accurate human pose estimation. \cite{savva2016pigraphs,savva2014scenegrok,hassan2021populating,hassan2021stochastic} generate human motion with interaction from the relationship between scene geometry and human body parts. \cite{monszpart2019imapper} further utilizes this relationship to recover interactions from videos. To explicitly use scene information to improve pose accuracy, \cite{hassan2019resolving} formulates two constraints in optimization to reduce interpenetration and encourage appropriate contact. \cite{zhang2021learning} also adopts the optimization-based approach and proposes a smoothness prior to improve motion quality. However, numerical optimization with soft constraints is hard to avoid artifacts like interpenetration, which is the main concern for human-scene reconstruction. In contrast, our method relies on a physics simulator~\cite{coumans2021} to provide hard physical constraints. With the network-based distribution prior, our method can obtain accurate terrain interactions via neural motion control.

\section{Method}\label{sec:method}
We propose a framework with a non-differentiable physics simulator~\cite{coumans2021} to capture physically plausible human motion. We first describe the representations of our kinematic and dynamical characters~(\cref{sec:Preliminaries}). Then, an interaction constraint is designed to obtain reference motion with appropriate contact information~(\cref{sec:kinematic}). In addition, we introduce a distribution prior trained with a novel two-branch structure for neural motion control~(\cref{sec:distribution}). Finally, we regress a distribution and sample satisfied target poses to track the estimated reference motion~(\cref{sec:sample}).

\subsection{Preliminaries}\label{sec:Preliminaries}
\noindent\textbf{Representation}. The kinematic motion is represented with SMPL model~\cite{loper2015smpl}. To represent different human shapes in the physics simulator, we design our physical character to have the same kinematic tree as SMPL. The bone length and link shape of the character can be directly obtained from the estimated SMPL parameters. We fix a few skeleton joints to have 57 DOFs. The state of character is denoted $\boldsymbol{s} = \left(\boldsymbol{q}, \dot{\boldsymbol{q}}\right)$, where $\boldsymbol{q}$ and $\dot{\boldsymbol{q}}$ are the pose and velocity, respectively. The details of the model can be found in the supplementary material.

\noindent\textbf{Sampling-based motion control}. We briefly review the sampling-based motion control approach~\cite{liu2010sampling} to promote understanding of our method. A kinematic pose $\tilde{\boldsymbol{q}}_{t}$ is used as a reference, and we wish the physical character to dynamically track the reference pose via PD-control~(Proportional Derivative). However, due to the inaccuracies of kinematic pose estimation and PD controller, the tracking always fails when directly applying the reference pose as the desired setpoint. The sampling algorithm samples a correction $\Delta \tilde{\boldsymbol{q}}_{t}$ for reference pose, thus employing the target pose $\overline{\boldsymbol{q}}_{t} = \tilde{\boldsymbol{q}}_{t} + \Delta \tilde{\boldsymbol{q}}_{t}$ can compensate the discrepancies. The quality of samples is evaluated by a loss function. By selecting the sample with the lowest loss, we can obtain the physically plausible motion. More details can be found in~\cite{liu2010sampling}.

\subsection{Reference motion estimation}\label{sec:kinematic}
The neural motion control requires reference motion with accurate ground contact to drive the physical character. To obtain the contact information, previous works~\cite{shimada2020physcap,Yu:2021:MovingCam} train a network to estimate a binary foot contact status. However, no sufficient data can be utilized for training in complex terrain scenarios~(\eg, stairs and uneven ground). We address the problem by incorporating an SDF-based interaction constraint in an optimization-based framework. 

Specifically, we optimize the latent code of pre-trained motion prior in~\cite{huang2021dynamic} to fit SMPL models to single-view 2D poses detected by AlphaPose~\cite{fang2017rmpe}. The overall formulation is:

\begin{equation}
    \underset{(\mathbf{z}, \mathcal{R}, \mathcal{T})_{1: T}, \beta}{\arg \min } \mathcal{L}=\mathcal{L}_{\text {data }}+\mathcal{L}_{\text {prior }}+\mathcal{L}_{\text {scene }},
\end{equation}
where $\mathbf{z}, \mathcal{R}, \mathcal{T}$ are the latent code, global rotation and translation for character in each frame. $\beta$ is the human shape parameter, and $T$ is the frame length. The data term is:
\begin{equation}
    \mathcal{L}_{\text {data}} =\sum_{t=1}^{T} \sigma_{t} \left\|\Pi \left(\tilde{\boldsymbol{j}}_{t}\right)-\mathbf{p}_{t}\right\|^2,\label{equ:data}
\end{equation}
where $\mathbf{p}$, $\sigma$ are 2D poses and their corresponding confidence. $\tilde{\boldsymbol{j}}$ is the model joint position. We further add the regularization term:
\begin{equation}
    \mathcal{L}_{\text{prior}} = \left\|\beta\right\|^{2} + \sum_{t=1}^{T} \left( \left\|\mathbf{z}_{t}\right\|^{2} + \left\|z_{t+1}-2 z_{t}+z_{t-1}\right\|^{2} \right).
\end{equation}

Due to the depth ambiguity, the recovered 3D human may float in the air or penetrate with the ground mesh with only the above constraints. With such reference motion, the simulated results are unnatural and incorrect. To reconstruct more accurate human-scene interactions from single-view videos, we generate a differentiable SDF of the scene mesh using~\cite{jiang2020coherent}. In the optimization, we follow~\cite{hassan2019resolving} to sample the SDF value for the pre-defined foot keypoints and use it to construct an objective function:

\begin{equation}
    \mathcal{L}_{\text {scene }} = \rho \| \operatorname{SDF}(\check{j}) \|^{2},
\end{equation}
where $\check{j}$ is the 3D positions of the keypoints and $\operatorname{SDF}$ is the sample operation. Our optimization has four stages. Since the proximate motion can be obtained in the first three stages, we only apply the interaction term to refine the ground contact in the last stage. To make our method to be compliant with airborne motions, we further apply a Geman-McClure error function $\rho$~\cite{geman1987statistical} to down-weight keypoints that are far from the scene mesh.

\begin{figure}
    \begin{center}
    \includegraphics[width=1\linewidth]{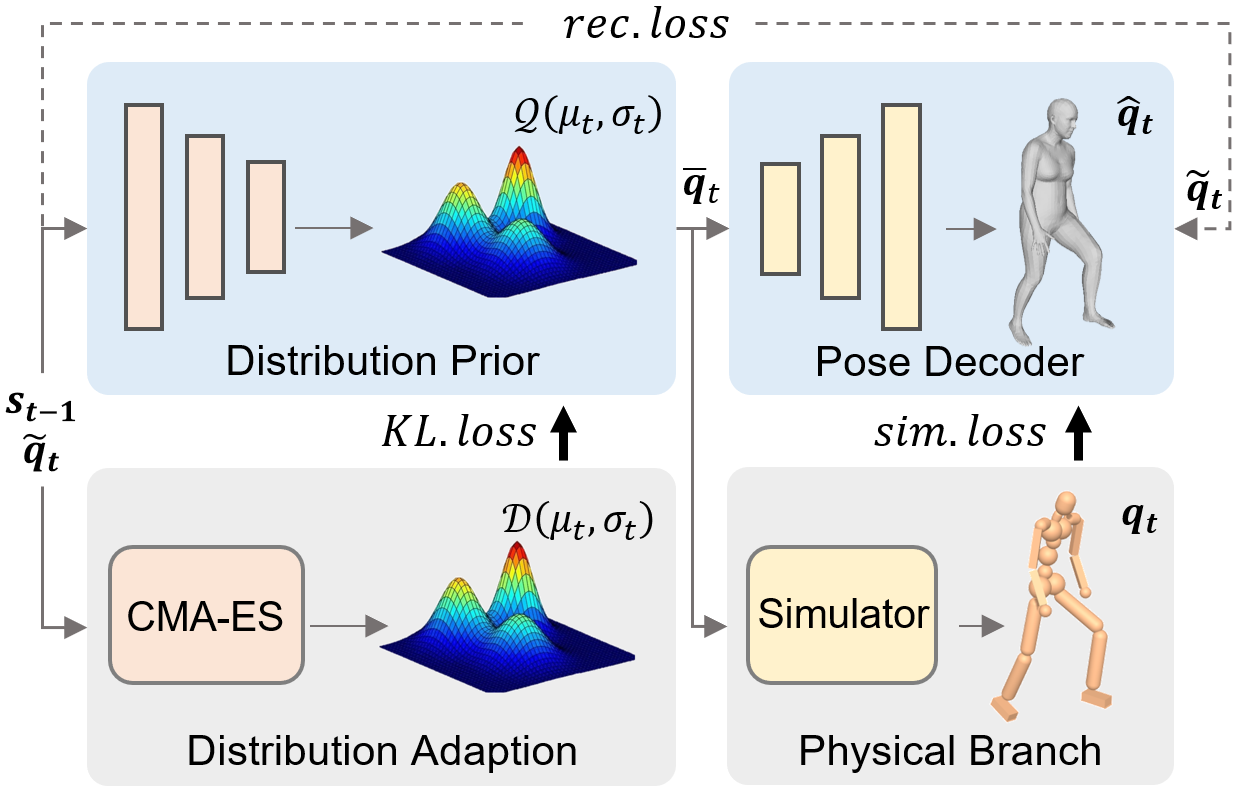}
    \end{center}
    \vspace{-6mm}
    \caption{Different from conventional approaches~(gray). We propose a two-branch decoder to avoid stochastic error from the CMA-ES method and train the distribution prior with real physical supervision. The non-differentiable physical branch simulates the sampled result, and the pose decoder intermediately employs the physical information to optimize the prior with simulation loss and reconstruction loss.}
    \label{fig:distribution}
    \vspace{-6mm}
    \end{figure}

\subsection{Distribution prior training}\label{sec:distribution}
It is essential to find an optimal target pose distribution to achieve physically plausible motion for sampling-based motion control. Previous works~\cite{liu2015improving} use $\left(\mu_{W}, \lambda\right)$-CMA-ES method~\cite{hansen2006cma} to realize the distribution adaption. However, the time-consuming operation and stochastic error of the adaption make it hard to be leveraged in motion capture for real-world applications. We propose to replace this operation and improve the performance with a network-based distribution prior. To train the network, a naive idea is to directly supervise the distribution using the CMA results. Given a pair of character state and reference pose, it seems that we can provide the supervisions by running CMA online before feeding the data into the network or preparing the pseudo supervision with CMA in advance. Actually, the two strategies are both infeasible in real implementation. For the same character state and reference pose, the CMA method obtains different distributions, resulting in network divergence and erroneous regression for online and offline strategies, respectively.

To solve this obstacle, we propose a two-branch decoder to assist training an accurate and generalized distribution encoder. As shown in the~\cref{fig:distribution}, we first pre-train the distribution encoder with the supervision from offline CMA. Since the network parameters trained with the inaccurate supervisions are incorrect, we then introduce a physical branch to verify the validity of the sampled target pose. Due to the non-differentiability of the simulator, we further design a pose decoder to intermediately employ physical supervision to train the distribution encoder.

Specifically, the KL divergence with pseudo ground-truth distributions is used to pre-train the encoder:
\begin{equation}
    \mathcal{L}_{kl}=KL(\mathcal{Q}(\Delta \tilde{\boldsymbol{q}}_{t} \mid \boldsymbol{s}_{t-1}, \tilde{\boldsymbol{q}}_t) \| \mathcal{D}(\mu_t, \sigma_t)),
\end{equation}
where $\mathcal{D}(\mu, \sigma)$ is the distribution prepared by $\left(\mu_{W}, \lambda\right)$-CMA-ES method and $\mathcal{Q}(\Delta \tilde{\boldsymbol{q}} \mid \boldsymbol{s}, \tilde{\boldsymbol{q}})$ is the estimated distribution. To improve the generalization ability, we sample correction of the reference pose from the estimated distribution, which is denoted as $\Delta \tilde{\boldsymbol{q}}_{t}$. Thus, the target pose is $\overline{\boldsymbol{q}}_{t} = \tilde{\boldsymbol{q}}_{t}+\Delta \tilde{\boldsymbol{q}}_{t}$.

To optimize the distribution encoder with real physical supervision, the sampled target pose is fed to the non-differentiable physics simulator to get the simulated pose. We design a pose decoder to imitate the physical branch by supervising it with the simulated pose. 
\begin{equation}
    \mathcal{L}_{sim}=\left\|\hat{\boldsymbol{q}}_{t}-\boldsymbol{q}_{t}\right\|^{2} + \left\|\hat{\boldsymbol{j}}_{t}-\boldsymbol{j}_{t}\right\|^{2},\label{equ:sim loss}
\end{equation}
where $\hat{\boldsymbol{q}}$, $\hat{\boldsymbol{j}}$ and $\boldsymbol{q}_{t}$, $\boldsymbol{j}$ are pose and joint positions of the estimated result and the simulated result, respectively. In addition, a reconstruction loss is applied to enforce optimal distribution encoding:
\begin{equation}
    \mathcal{L}_{rec}=\left\|\hat{\boldsymbol{q}}_{t}-\tilde{\boldsymbol{q}}_{t}\right\|^{2} + \left\|\hat{\boldsymbol{j}}_{t}-\tilde{\boldsymbol{j}}_{t}\right\|^{2}.\label{equ:rec loss}
\end{equation}
With the pose decoder, the encoder can gradually encode valid distribution to sample effective poses in the simulator. We further add a regularization term to ensure the network will not be easily overfitted:
\begin{equation}
    \mathcal{L}_{reg}=\|\phi\|_{2}^{2}.
\end{equation}

We reduce the weight of KL loss when training with the two-branch decoder. The overall loss function is:
\begin{equation}
    \mathcal{L}_{dist}=\mathcal{L}_{sim}+\mathcal{L}_{rec}+\lambda\mathcal{L}_{kl}+\mathcal{L}_{reg}.
\end{equation}

The $\lambda$ is 0.2 in our experiments. When the training is finished, the encoder is utilized to construct a neural motion capture framework in \cref{sec:sample}.

\subsection{Motion capture with neural motion control}\label{sec:sample}
With the trained distribution prior, we then capture human motion by tracking the kinematic reference motion by a sampling strategy. As shown in~\cref{fig:pipeline}, the reference pose and the current state of character are first fed into the prior to encode target pose distribution. Then, we sample target poses and simulate them in the simulator. The quality of each sample is evaluated with character-level and image-level loss functions. The sample with the lowest loss will be adopted for the next frame. Since the reference motions from uneven terrains are noisy, we design several loss functions to evaluate sample quality.

The loss between simulated pose and reference pose is first used to measure the pose and joint position consistency. 
\begin{equation}
    \mathcal{L}_{tra}=\left\|\boldsymbol{q}_{t}-\tilde{\boldsymbol{q}}_{t}\right\|^{2} + \left\|\boldsymbol{j}_{t}-\tilde{\boldsymbol{j}}_{t}\right\|^{2}.
\end{equation}
We find that the dynamical state of the character is critical for physics-based motion capture. We then introduce a dynamical loss to evaluate the velocity consistency:
\begin{equation}
    \mathcal{L}_{dyn}=\left\|\dot{\boldsymbol{q}}_{t}-\dot{\tilde{\boldsymbol{q}}}_{t}\right\|^{2} + \left\|\dot{\boldsymbol{j}}_{t}-\dot{\tilde{\boldsymbol{j}}}_{t}\right\|^{2},\label{equ:dyn loss}
\end{equation}
where $\dot{\boldsymbol{q}}$ and $\dot{\boldsymbol{j}}$ are joint angular velocity and linear velocity, respectively. To let the physical character keep balance, we follow~\cite{liu2010sampling} to add a balance term to adjust CoM~(Center of Mass):

\begin{equation}
    \mathcal{L}_{ban}= \sum_{m=0}^{M} \left\|\boldsymbol{d}_{t}^{m}-\tilde{\boldsymbol{d}}_{t}^{m}\right\|^{2} + \left\|\dot{\boldsymbol{j}}_{t}^{CoM}-\dot{\tilde{\boldsymbol{j}}}_{t}^{CoM}\right\|^{2},
\end{equation}
where, $\boldsymbol{d}^{m}=(\boldsymbol{j}^{m}-\boldsymbol{j}^{CoM})|_{z=0}$, which denotes the planar vector from end-effector $m$ to CoM. The $\dot{\boldsymbol{j}}^{CoM}$ is the linear velocity of CoM and $M$ is number of end-effectors.

Different from DRL, we can directly use image features to evaluate the quality of the sample. With 2D pose and corresponding confidence, the image-level loss makes our method more robust to occlusion scenarios:
\begin{equation}
    \mathcal{L}_{reproj} =\sigma \left\|\Pi \left(\boldsymbol{j}_{t}\right)-\mathbf{p}_{t}\right\|^2.
\end{equation}

The overall loss function for the sampling procedure is:
\begin{equation}
    \mathcal{L}_{sam}=\mathcal{L}_{tra}+\mathcal{L}_{dyn}+\mathcal{L}_{ban}+\mathcal{L}_{reproj}.
\end{equation}
Finally, the sample with the lowest loss in each frame consists of a complete physically plausible human motion. 

\section{Experiments}\label{sec:experiment}
In this section, we conduct several qualitative and quantitative experiments to demonstrate the effectiveness of our method. We first introduce the implementation details and datasets in~\cref{sec:Metrics} and~\cref{sec:datasets}. Then, the comparisons with the state-of-the-arts are shown in~\cref{sec:comparison}. Finally, ablation studies in~\cref{sec:ablation} are conducted to evaluate key components.

\begin{figure*}
    \begin{center}
    \includegraphics[width=1.0\linewidth]{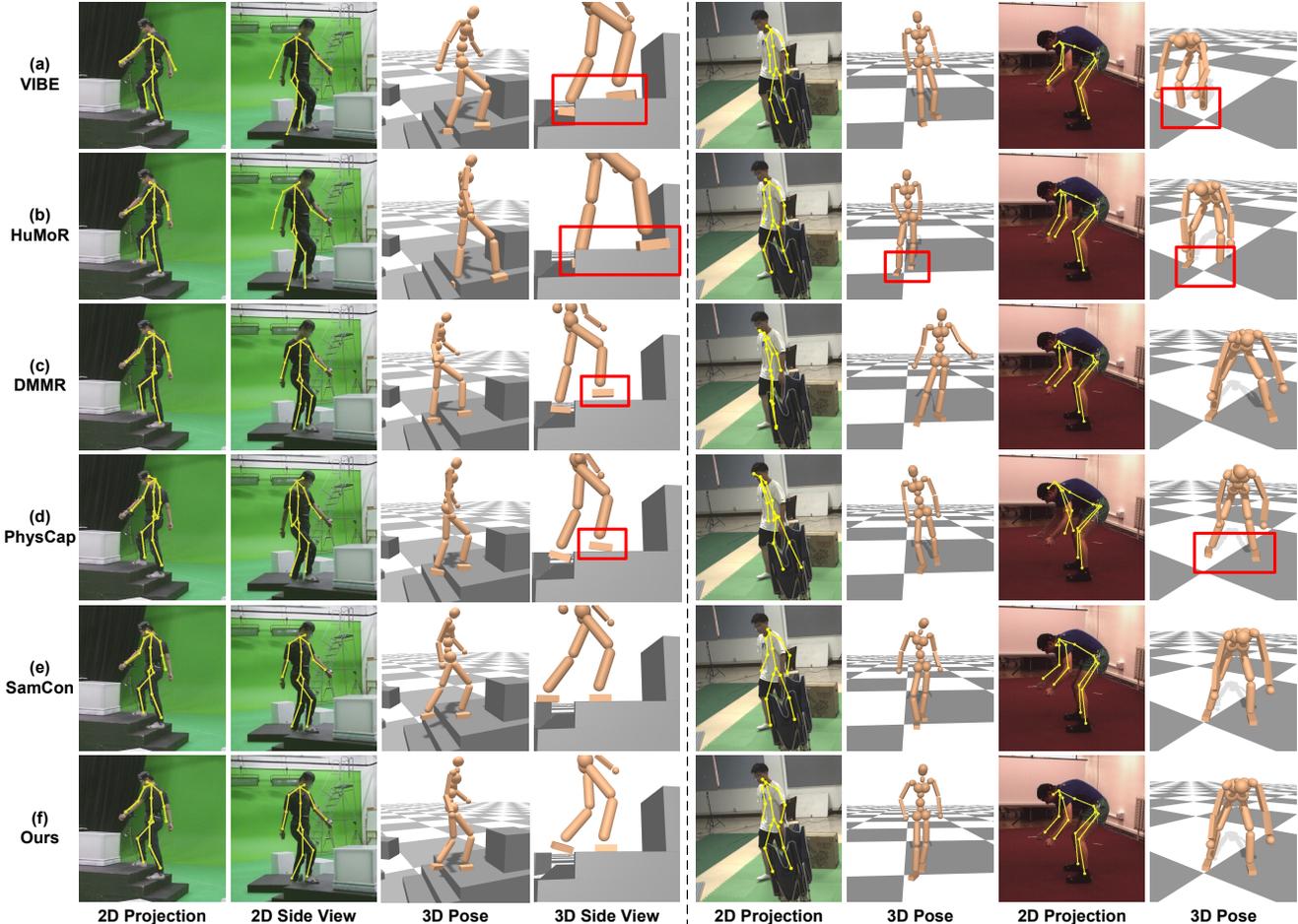}
    \end{center}
    \vspace{-8mm}
    \caption{Qualitative comparison with other methods. For a fair comparison, we represent all results using our character with corresponding shape variations. The results show that our method can obtain physically plausible and natural human motion from monocular RGB videos.}
    \label{fig:qualitative_comparision}
    \vspace{-6mm}
    \end{figure*}

\subsection{Metrics}\label{sec:Metrics}
The common metrics of the Mean Per Joint Position Error~(MPJPE) and the MPJPE after rigid alignment of the prediction with ground truth using Procrustes Analysis (MPJPE-PA) are used to evaluate joint accuracy. To evaluate physical plausibility, we use the metrics proposed in~\cite{shimada2020physcap} and~\cite{xie2021physics} to   measure motion jitter and foot contact. $e_{S}$ is the difference in joint velocity magnitude between the ground truth motion and the predicted motion. $e_{S}$ and its standard deviation $\sigma_{S}$ are used to assess motion smoothness. $e_{f, z}$ is the foot position error on z-axis. We adopt this metric to evaluate foot floating artifacts. More details can be found in their original paper.

\subsection{Datasets}\label{sec:datasets}
\textbf{Human3.6M}~\cite{h36m_pami} is a large-scale dataset, which consists of 3.6 million 3D human poses and corresponding images. Following previous work~\cite{yuan2021simpoe}, we train our model on 5 subjects~(S1,S5,S6,S7,S8), and test on the other subjects~(S9,S11) with 25Hz.

\textbf{GPA}~\cite{wang2019geometric} is a 3D human dataset with both human-scene interactions and ground-truth scene geometries. It utilizes a commercial motion capture system to collect data. The sequence \textit{0, 34, 52} are used to test, and the rest are served as training data. With the scene geometries, we verify the performance of our method on more complex terrains.

\textbf{3DOH}~\cite{zhang2020object} is the first dataset to handle the object-occluded human body estimation problem, which contains 3D motions in occluded-scenarios. We use the sequence \textit{13, 27, 29} in this dataset to evaluate our method on occlusion cases.

\textbf{GTA-IM}~\cite{cao2020long}. Since there are limited ground-truth terrain data, we use this synthetic dataset as additional human-scene interaction cases. The scene meshes are recovered from the depth map. We conduct qualitative experiments on this dataset.

\begin{table}
    \begin{center}
        \resizebox{1.0\linewidth}{!}{
            \begin{tabular}{l|c c c c c}
            \noalign{\hrule height 1.5pt}
            \begin{tabular}[l]{l}\multirow{1}{*}{Method}\end{tabular}
                &MPJPE  &PA-MPJPE &$e_{S}$ &$\sigma_{S}$ &$e_{f, z}$ \\
            \noalign{\hrule height 1pt}
            $^*$HuMoR~\cite{rempe2021humor}   &97.5 &68.5 &24.2 &25.9 &43.2  \\
            $^*$DMMR~\cite{huang2021dynamic}   &96.0  &67.4  &\textbf{14.4}  &\textbf{12.6} &48.6    \\
            $^*$VIBE~\cite{kocabas2020vibe}   &\textbf{65.9}  &\textbf{41.5}  &25.5  &25.7 &\textbf{34.0}       \\
            \hline \hline
            EgoPose~\cite{yuan2019ego}   &130.3  &79.2  &--  &-- &--   \\
            PhysCap~\cite{shimada2020physcap}   &97.4  &65.1  &7.2  &6.9 &--    \\
            SamCon~\cite{liu2015improving}   &78.4   &63.2   &4.0   &4.3 &20.4     \\
            NeuralPhysCap~\cite{shimada2021neural}   &76.5 &58.2  &4.5  &6.9 &--    \\
            Xie~\etal~\cite{xie2021physics}   &68.1 &--  &4.0  &\textbf{1.3} &18.9    \\
            SimPoE~\cite{yuan2021simpoe}   &\textbf{56.7}  &\textbf{41.6}  &--  &-- &--    \\
            
            \textbf{Ours}   &72.5  &54.6  &\textbf{3.8}  &2.4 &\textbf{14.4}    \\
            \noalign{\hrule height 1.5pt}
            \end{tabular}
        }
    \vspace{-3mm}
    \caption{Comparisons with state-of-the-art methods on Human3.6M dataset. Our method achieves good performance in physical plausibility and motion smoothness. $^*$ denotes the kinematics-based method.}
    \label{tab:h36m}
    \end{center}
    \vspace{-11mm}
    \end{table}

\begin{table}
    \begin{center}
        \resizebox{1.0\linewidth}{!}{
            \begin{tabular}{l|c c c|c c c  c c c c}
            \noalign{\hrule height 1.5pt}
            \begin{tabular}[l]{l}\multirow{2}{*}{Method}\end{tabular}

            &\multicolumn{3}{c|}{\textit{3DOH}} &\multicolumn{3}{c}{\textit{GPA}}\\
            & MPJPE & PA-MPJPE & $e_{S}$ & MPJPE & PA-MPJPE & $e_{f, z}$  \\
            \noalign{\hrule height 1pt}
            $^*$DMMR~\cite{huang2021dynamic} &102.9  &65.8  &\textbf{16.2}  &\textbf{107.0} &87.4 &\textbf{32.8}    \\
            $^*$VIBE~\cite{kocabas2020vibe} &\textbf{98.1}  &61.8  &26.5  &114.3  &\textbf{80.6} &36.4  \\
            $^*$HuMoR~\cite{rempe2021humor} &105.1 &\textbf{60.6} &21.9 &117.2 &86.3 &58.7 \\
            \hline \hline
            SamCon~\cite{liu2015improving}   &102.4   &95.4   &9.7   &104.7 & 87.1 &28.3     \\
            PhysCap~\cite{shimada2020physcap} &107.8 &93.3  &12.2  &103.4  &91.2 &36.1  \\
            \textbf{Ours}                 &\textbf{93.4} &\textbf{86.7} &\textbf{9.2} &\textbf{94.8} &\textbf{80.3} &\textbf{21.2} \\
            \noalign{\hrule height 1.5pt}
            \end{tabular}
        }
    \end{center}
    \vspace{-6mm}
    \caption{Quantitative comparison on 3DOH and GPA dataset. Our method achieves state-of-the-art in complex terrain scenarios and occlusion cases. $^*$ denotes the kinematics-based method.}
    \label{tab:3doh_gpa}
    \vspace{-7mm}
    \end{table}

\begin{figure*}
    \begin{center}
    \includegraphics[width=1.0\linewidth]{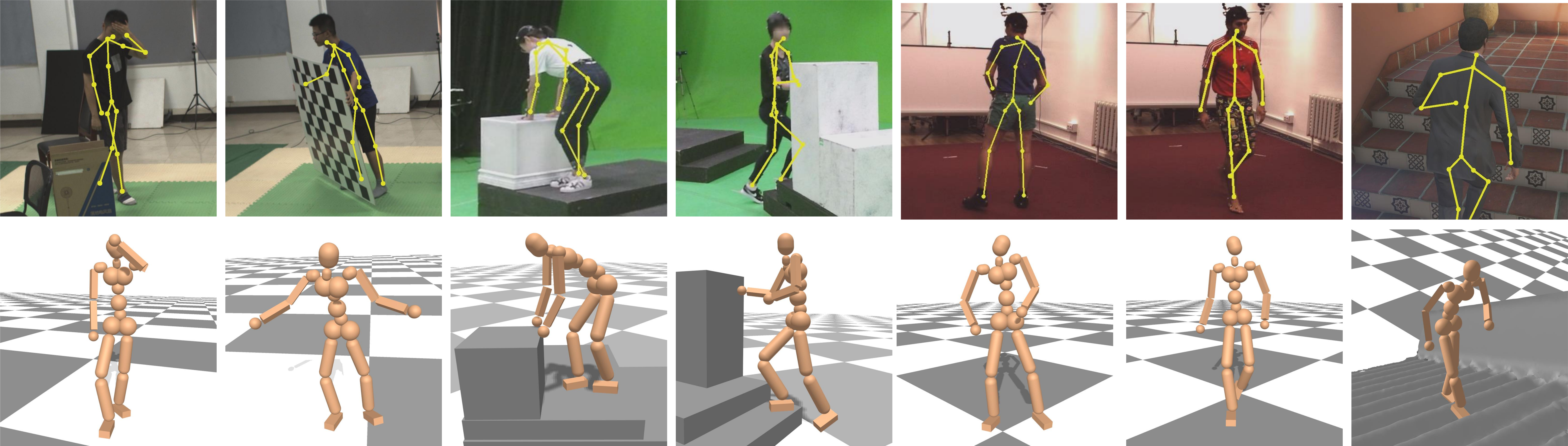}
    \end{center}
    \vspace{-6mm}
    \caption{Our method is general to different terrain interactions, human shape variations, and diverse behaviors.}
    \label{fig:qualitative_results}
    \vspace{-6mm}
    \end{figure*}

\subsection{Comparison to state-of-the-art methods}\label{sec:comparison}
There are several kinematic and dynamical approaches that report results on Human3.6M datasets. As shown in~\cref{tab:h36m}, we first evaluated our method on this dataset to demonstrate that our neural motion control works well on flat ground. \cite{kocabas2020vibe,huang2021dynamic,rempe2021humor} are recent works to estimate kinematic SMPL parameters. Although the explicit dynamics of the human model are not considered, \cite{huang2021dynamic,rempe2021humor} learn implicit dynamics via VAE and improve physical plausibility by using prior knowledge. The rest methods in~\cref{tab:h36m} are dynamics-based methods. Specifically, SamCon~\cite{liu2015improving} is designed for animation. We used this method to track our kinematic motion and adopted it as a baseline to compare among sampling-based methods.

In~\cref{tab:h36m}, we found that VIBE achieves the best performance in terms of PA-MPJPE. It relies on a GRU-based network to build correspondences among different frames. However, directly regressing kinematic SMPL parameters causes the largest smoothness error and results in visually noticeable motion jitter. Furthermore, VIBE shows a severe penetration with the ground in~\cref{fig:qualitative_comparision}. Due to model discrepancies between the motion capture subject and the physical character, the joint position error for dynamics-based methods is higher than kinematics-based approaches. SimPoE~\cite{yuan2021simpoe} utilizes a model with a similar shape as Human3.6M subjects and get comparable results to VIBE. However, for different subjects with the variation of body proportion and shape, this method requires to re-train the policy. Benefited from the proposed target pose distribution prior, our method can adapt to shape variation. Thus, we can update the bone length of the physical character model with the estimated human shape and directly use it to capture human motion from images. Our method also obtained smooth motion and achieves state-of-the-art in terms of $e_S$.

We then compared our method to others on the 3DOH dataset. It is tricky to obtain accurate reference motion for occlusion cases. As shown in the 5th column of~\cref{fig:qualitative_comparision}, the inaccurate reference motion will result in a large deviation between 3D pose and image observation for other physics-based methods. However, due to the image-level loss, our method got more accurate results. Moreover, SamCon also based on a sampling approach to get human motion. The results in~\cref{tab:3doh_gpa} and \cref{fig:qualitative_comparision} show that our network-based distribution prior can get more appropriate distribution and then produce natural and precise motion.

On the GPA dataset, we evaluated our method with complex terrains. The interactions with objects and terrains impose great difficulty for kinematics-based methods. The estimated poses float on the air or penetrate with the scene mesh for their methods~(\cref{fig:qualitative_comparision}). Since PhysCap uses a numerical optimization framework with soft physical constraints to capture human motion, the results also show physical artifacts. The qualitative and quantitative results on GPA dataset in~\cref{fig:qualitative_comparision} and~\cref{tab:3doh_gpa} show that neural motion control is more proper for contact-rich scenarios.

\begin{figure*}
    \begin{center}
    \includegraphics[width=1.0\linewidth]{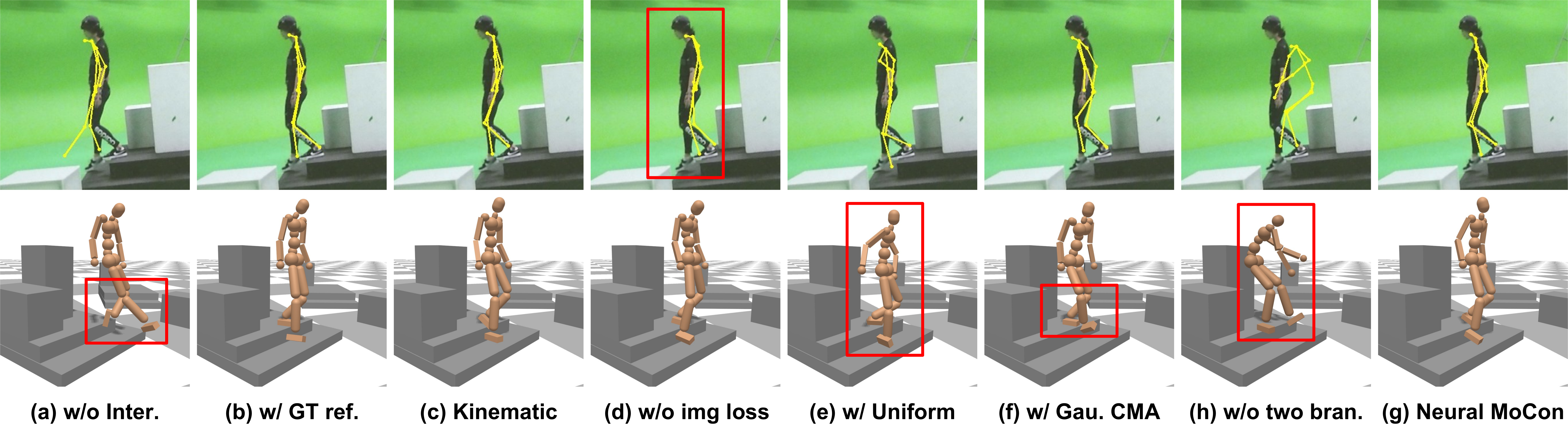}
    \end{center}
    \vspace{-7mm}
    \caption{Ablation on different components. (a, d, h) are the results of our method that removes interaction constraint, image-level loss, and two-branch decoder, respectively. (e, f) replace the distribution prior with uniform distribution and gaussian distribution. (c) is the kinematic result from our optimization and (b) is the simulated result with ground-truth reference motion.}
    \label{fig:abaltion}
    \vspace{-5mm}
    \end{figure*}

\subsection{Ablation studies}\label{sec:ablation}
\textbf{Two-branch decoder}. As mentioned before, directly supervising the distribution encoder without two-branch decoder will result in erroneous regression. In~\cref{fig:abaltion} and~\cref{tab:ablation}, we conducted comparisons between the distribution prior trained with and without the two-branch decoder. Without the decoder, the encoder can not regress correct distribution to sample a valid target pose, thus causing an unsatisfied simulated pose. The quantitative results in~\cref{tab:ablation} show that the two-branch decoder induces major improvement and demonstrate that it is the most important component for our method. 

\textbf{Distribution prior}. We compared different methods of distribution generation to verify the superiority of our distribution prior. We first replaced the distribution encoder with uniform distribution with a pre-defined range. The results in~\cref{fig:abaltion} show that it can not generalize to a large variety motion types. As shown in~\cref{tab:ablation}, since there is a stochastic error for the CMA method, the gaussian distribution with CMA adaption is inferior to the distribution encoder.

\begin{figure}
    \begin{center}
    \includegraphics[width=0.9\linewidth]{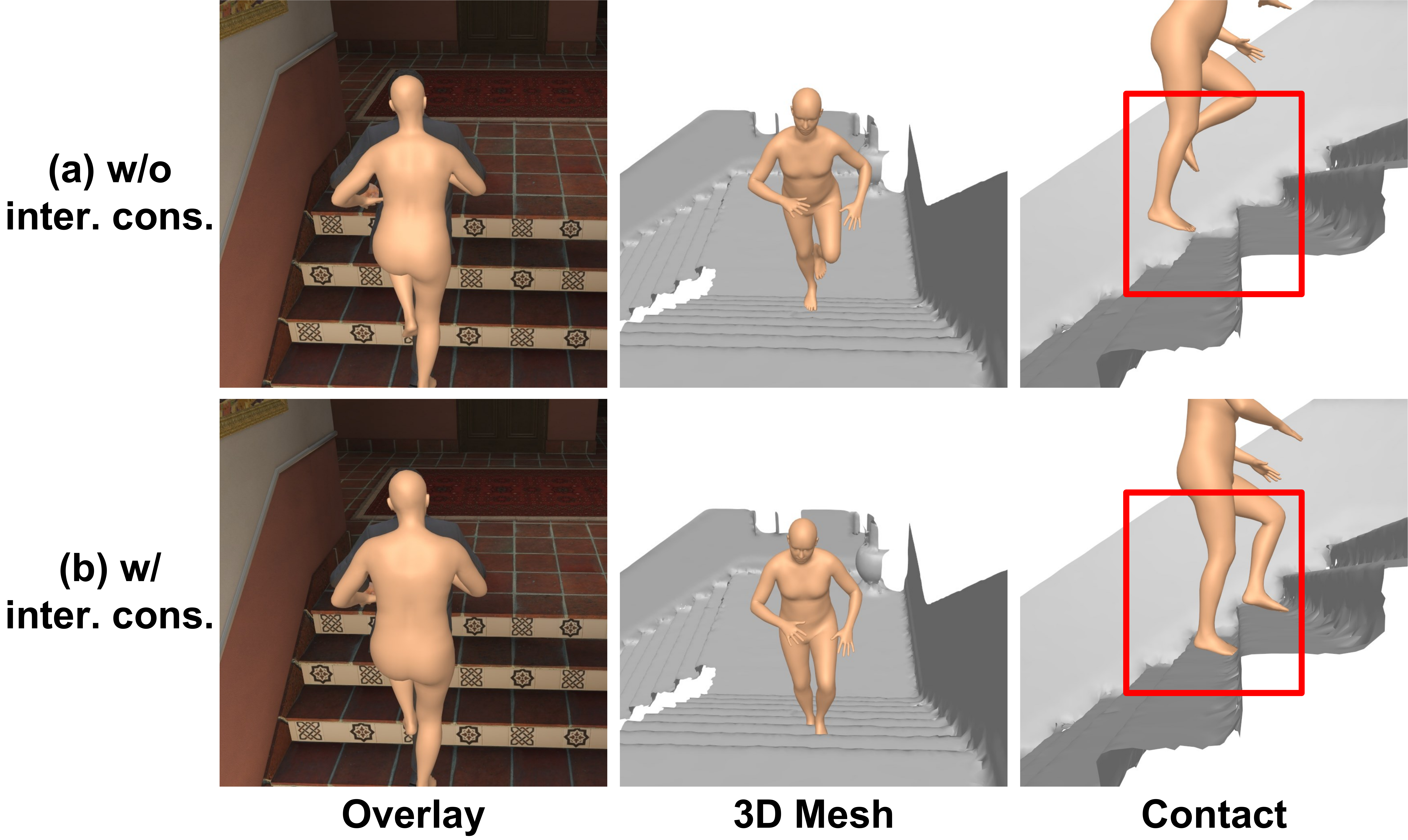}
    \end{center}
    \vspace{-7mm}
    \caption{The kinematic reference motion obtained with and without interaction constraint on complex terrain.}
    \label{fig:interloss}
    \vspace{-4mm}
    \end{figure}

\textbf{Interaction constraint}. We further conducted several experiments to illustrate the necessity of the interaction constraint. Due to the visual ambiguity, it is difficult to reconstruct accurate human-scene interactions with complex terrains~(\cref{fig:interloss}). In \cref{tab:ablation}, the optimization with interaction constraint gets more accurate foot position on GPA dataset. In addition, an inaccurate contact seriously affects the performance of sampling-based motion control. \cref{fig:abaltion}~(a) shows a reference pose floating on the air can trigger improper simulated pose. The gap between the results of the method with and without this constraint on GPA is greater than that on 3DOH in~\cref{tab:ablation}, which proves its importance for motion capture on complex terrains.

\begin{table}
    \begin{center}
        \resizebox{1.0\linewidth}{!}{
            \begin{tabular}{l|c c c|c c c  c c c c}
            \noalign{\hrule height 1.5pt}
            \begin{tabular}[l]{l}\multirow{2}{*}{Method}\end{tabular}

            &\multicolumn{3}{c|}{\textit{GPA}} &\multicolumn{2}{c}{\textit{3DOH}}\\
            & MPJPE & PA-MPJPE & $e_{f, z}$ & MPJPE & $e_{S}$  \\
            \noalign{\hrule height 1pt}
            $^*$DMMR~\cite{huang2021dynamic} &107.0  &87.4  &32.8  &102.9 &16.2    \\
            $^*$Kinematic        &106.2  &87.2   &27.3  &94.4 &16.5    \\
            \hline \hline
            w/o two-branch       &142.2  &126.7  &28.4  &136.8  &13.2  \\
            w/ Uniform Dist.     &136.6  &119.1  &29.6  &142.1  &10.3  \\
            w/o Inter. Cons.     &116.4  &109.4  &24.3  &93.4   &9.4 \\
            w/ Gaussian CMA      &103.9   &84.4   &23.5  &95.4   &9.8  \\
            w/o image-level loss &95.8   &84.4   &21.3  &96.3   &9.7  \\
            w/ GT reference      &93.6   &80.0   &17.3  &89.6   &9.2  \\
            Neural MoCon         &94.8   &80.3   &21.2  &93.4   &9.2 \\
            \noalign{\hrule height 1.5pt}
            \end{tabular}
        }
    \end{center}
    \vspace{-6mm}
    \caption{Quantitative results of ablation studies. w/o denotes to remove corresponding component of our method. w/ Uniform Dist. and w/ Gaussian CMA indicate to replace distribution prior with uniform and gaussian distribution. w/ GT reference uses ground-truth reference motion for neural motion control. $^*$ denotes the kinematics-based method.}
    \label{tab:ablation}
    \vspace{-7mm}
    \end{table}

\section{Limitation and future work}\label{sec:limitation}

Although our method can obtain physically plausible human motion via neural motion control, there are some limitations for the current implementation. First, the discrepancy between the geometric primitives of our character and the real human body makes our method unable to reconstruct accurate body contact~(\eg, Lying on the sofa). To solve this problem, building a more delicate character model like~\cite{yuan2021simpoe} may be a feasible approach. Second, the cumulative error of an undesirable sample may result in failure to sample a long sequence. Future work can integrate long-term temporal information in the sampling. Finally, due to a lack of ground-truth terrain data, we can only evaluate our method on similar interactions like stairs for motion capture tasks. Therefore, to build a large-scale human-scene interaction dataset for human motion capture in complex scenarios is also worthwhile.

Among Neural MoCon, DRL-based methods, and traditional sampling-based motion control, DRL can obtain highly accurate results for a specific task, and sampling control is more general to unknown scenarios. Neural MoCon is in between these two typical technical approaches. To combine the accuracy of DRL and the generalization ability of sampling control may be a potential direction to promote future physics-based motion capture.

\vspace{-2mm}
\section{Conclusion}\label{sec:conclusion}
In this paper, we propose a framework to capture physically plausible human motion with complex terrain interactions, human shape variations, and diverse behaviors. We first introduce an interaction constraint based on SDF in optimization to estimate accurate human-scene contact. Then, a novel two-branch decoder is designed to train a distribution prior with real physical supervision. With the trained prior and the estimated reference motion, several loss functions are used to select a satisfied sample to consist of a complete human motion. The proposed method has better generalization ability than DRL-based methods and gets more accurate results than conventional sampling-based motion control.

{\small
\bibliographystyle{ieee_fullname}
\bibliography{egbib}
}

\appendix
\begin{center}
\section*{Supplementary Material}
\end{center}

In the supplementary material, we first introduce implementation details of our method for reproducing the experimental results~(\cref{sec:review}). An additional evaluation on different metric is provided~(\cref{sec:Success}). Then, more results on different datasets are shown to demonstrate the performance of our method~(\cref{sec:experiments}). We further discuss the results in the supplementary video~(\url{https://www.bilibili.com/video/BV1W94y1f7ht})~(\cref{sec:video}). Finally, the superiority of the proposed neural motion control is discussed~(\cref{sec:why}). 

\vspace{2mm}

\section{Implementation details}\label{sec:review}
We adopt PyBullet~\cite{coumans2021} as simulator. The control frequency is 240HZ, and the coefficient of friction is 0.9. Since the frame rate among videos is different, we apply linear interpolation on the estimated motion between two frames to obtain reference pose and velocity. The frequency of sampling is 30HZ. To train the distribution prior, we implement the neural network based on PyTorch~\cite{paszke2019pytorch}. The distribution encoder and the pose decoder have six and four fully-connected layers, respectively, with batch normalization and LeakyReLU~\cite{maas2013rectifier} activation function. The AdamW~\cite{loshchilov2017decoupled} optimizer with a learning rate of 0.0001 is used to train the network. On a desktop with an Intel(R) Core(TM) i9-11900F CPU and a GPU of NVIDIA GeForce RTX 3090, one sample takes about 0.0002s without any implementation acceleration strategy. We sample 1000 samples for a target pose and save 20 samples as the start state for the next target pose.

\subsection{Physical character creation}\label{sec:character}
\begin{figure}
    \begin{center}
    \includegraphics[width=1.0\linewidth]{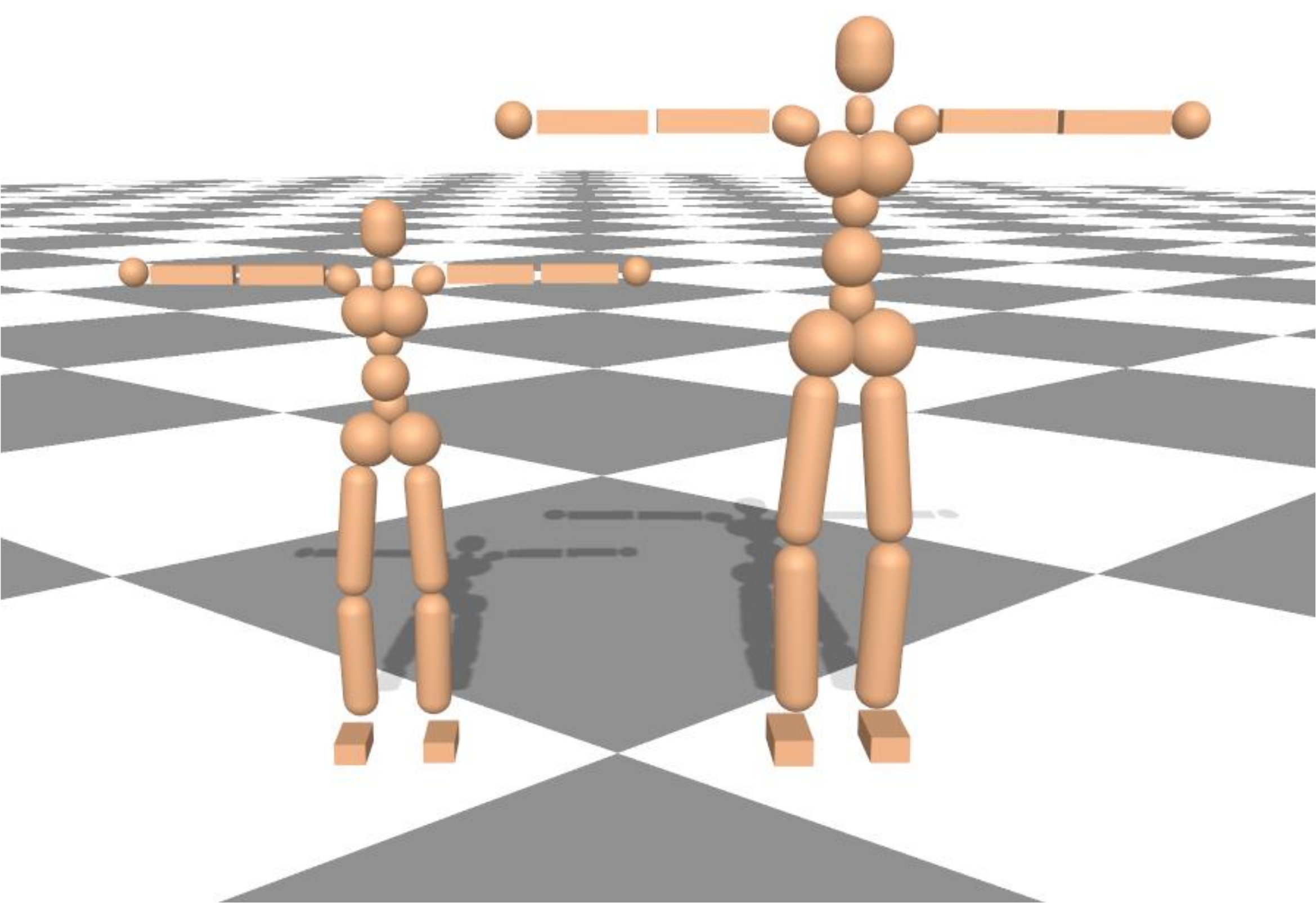}
    \end{center}
    \vspace{-6mm}
    \caption{The physical character with different body shapes. The character has the same joint positions as its corresponding SMPL model.}
    \label{fig:charactershape}
    \vspace{-7mm}
    \end{figure}

In this section, we explain the details of physical character creation with different body shape variations. To represent the kinematic and dynamical model in a unified framework, we design the kinematic tree of the physical character to be the same as the SMPL~\cite{loper2015smpl}. According to the estimated SMPL shape parameters, we automatically generate a new character. With the joint regressor in SMPL, we obtain the length of each bone of the estimated SMPL model in T-pose. Since the bones in symmetrical parts have minor difference, we calculate the average length and correct the rotation for each bone, and build the skeleton based on parent-child relationship. Further, we determine the link shape with the created skeleton. The physical characters with different shapes are shown in~\cref{fig:charactershape}. In addition, we do not control the hand and foot, so that these joints are fixed. Since the control parameters are dramatically affected by mass, all characters in our experiments have the same mass. The details of the character model are described in~\cref{tab:control}.

\subsection{CMA-ES}\label{sec:cmaes}
The CMA-ES~(covariance matrix adaptation evolution strategy)~\cite{hansen2006cma} is a black-box optimization method. We implement this algorithm with~\cite{Hansen16a} to prepare pseudo ground-truth for distribution prior training. The mean and the variance have the same dimension as target pose, which is 51. The number of maximum resampling is 100 and the population size is 6 in our experiments. The distribution evolves 30 generations for a given character state and reference pose. To get more natural motion, we limit the sampling bounds, which is shown in~\cref{tab:bounds}.

\begin{table*}
    \begin{center}
        \resizebox{1.0\linewidth}{!}{
            \begin{tabular}{l c c c c c c  c c c c c c c}
            \noalign{\hrule height 1.5pt}
            \begin{tabular}[l]{l}\multirow{2}{*}{Joint}\end{tabular}
            
            & \multirow{2}{*}{Type} & \multirow{2}{*}{Geometry} & \multirow{2}{*}{Mass} & \multirow{2}{*}{Num} & \multirow{2}{*}{Kp} & \multirow{2}{*}{Kd} & \multirow{2}{*}{Force Limit} & \multirow{2}{*}{Inertia(xx)} & \multirow{2}{*}{Inertia(xy)} & \multirow{2}{*}{Inertia(xz)} & \multirow{2}{*}{Inertia(yy)} & \multirow{2}{*}{Inertia(yz)} & \multirow{2}{*}{Inertia(zz)} \\

            &\multicolumn{3}{c}{\textit{}} \\

            \noalign{\hrule height 1pt}
            Lower Neck & revolute & capsule & 0.5 & 1 & 200  & 20 & 100 & 0.001 & 0.0 & 0.0 & 0.001 & 0.0 & 0.001\\
            Upper Neck & revolute & capsule & 3.0 & 1 & 200  & 20 & 100 & 0.001 & 0.0 & 0.0 & 0.001 & 0.0 & 0.001\\
            Chest & revolute & sphere & 8.0 & 1 & 500  & 50 & 300 & 0.001 & 0.0 & 0.0 & 0.001 & 0.0 & 0.001\\
            Lower Back & revolute & sphere & 5.0 & 1 & 500  & 50 & 300 & 0.001 & 0.0 & 0.0 & 0.001 & 0.0 & 0.001\\
            Upper Back & revolute & sphere & 5.0 & 1 & 500  & 50 & 300 & 0.001 & 0.0 & 0.0 & 0.001 & 0.0 & 0.001\\
            Clavicle & revolute & capsule & 1.0 & 2 & 400  & 40 & 200 & 0.001 & 0.0 & 0.0 & 0.001 & 0.0 & 0.001\\
            Shoulder & revolute & box & 2.0 & 2 & 400  & 40 & 200 & 0.001 & 0.0 & 0.0 & 0.001 & 0.0 & 0.001\\
            Elbow & revolute & box & 1.0 & 2 & 300  & 30 & 150 & 0.001 & 0.0 & 0.0 & 0.001 & 0.0 & 0.001\\
            Wrist & fixed & sphere & 0.5 & 2 & --  & -- & -- & 0.001 & 0.0 & 0.0 & 0.001 & 0.0 & 0.001\\
            Hip & revolute & capsule & 5.0 & 2 & 500  & 50 & 300 & 0.001 & 0.0 & 0.0 & 0.001 & 0.0 & 0.001\\
            Knee & revolute & capsule & 3.0 & 2 & 400  & 40 & 200 & 0.001 & 0.0 & 0.0 & 0.001 & 0.0 & 0.001\\
            Ankle & revolute & box & 1.0 & 2 & 300 & 30 & 100 & 0.001 & 0.0 & 0.0 & 0.001 & 0.0 & 0.001\\
            \noalign{\hrule height 1.5pt}
            \end{tabular}
        }
    \end{center}
    \vspace{-4mm}
    \caption{Control parameters and joint information. Inertia~(ij) represents the link inertia coefficient between the i-axis and j-axis.}
    \label{tab:control}
    \vspace{-2mm}
\end{table*}

\begin{table}
    \begin{center}
        \resizebox{1.0\linewidth}{!}{
            \begin{tabular}{l c| l c c c c  c c }
            \noalign{\hrule height 1.5pt}
            \begin{tabular}[l]{l}\multirow{2}{*}{Character Property}\end{tabular}
            
            & \multirow{2}{*}{Value} & \multirow{2}{*}{Simulator Property} & \multirow{2}{*}{Value} \\ 

            &\multicolumn{1}{c|}{\textit{}} \\

            \noalign{\hrule height 1pt}
   
            Joints  & 19  &Gravity  &9.81 \\
            Movable Joints   & 17  &Time Step &1/240.0  \\
            Fixed Joints  & 2  &NumSolverIterations &10   \\
            Links   & 19  &NumSubSteps &2  \\
            Total Mass (kg)     & 53.5  &   \\
            Degrees of Freedom     & 57  &   \\
            Lateral Friction Coefficient  & 0.9 & \\
            Rolling Friction Coefficient  & 0.3 & \\
            Restitution Coefficient  & 0.0 & \\
           
            \noalign{\hrule height 1.5pt}
            \end{tabular}
        }
    \end{center}
    \vspace{-6mm}
    \caption{Properties of the physical character and the physics simulator.}
    \label{tab:property}
    \vspace{-2mm}
    \end{table}

\begin{figure*}
    \begin{center}
    \includegraphics[width=1.0\linewidth]{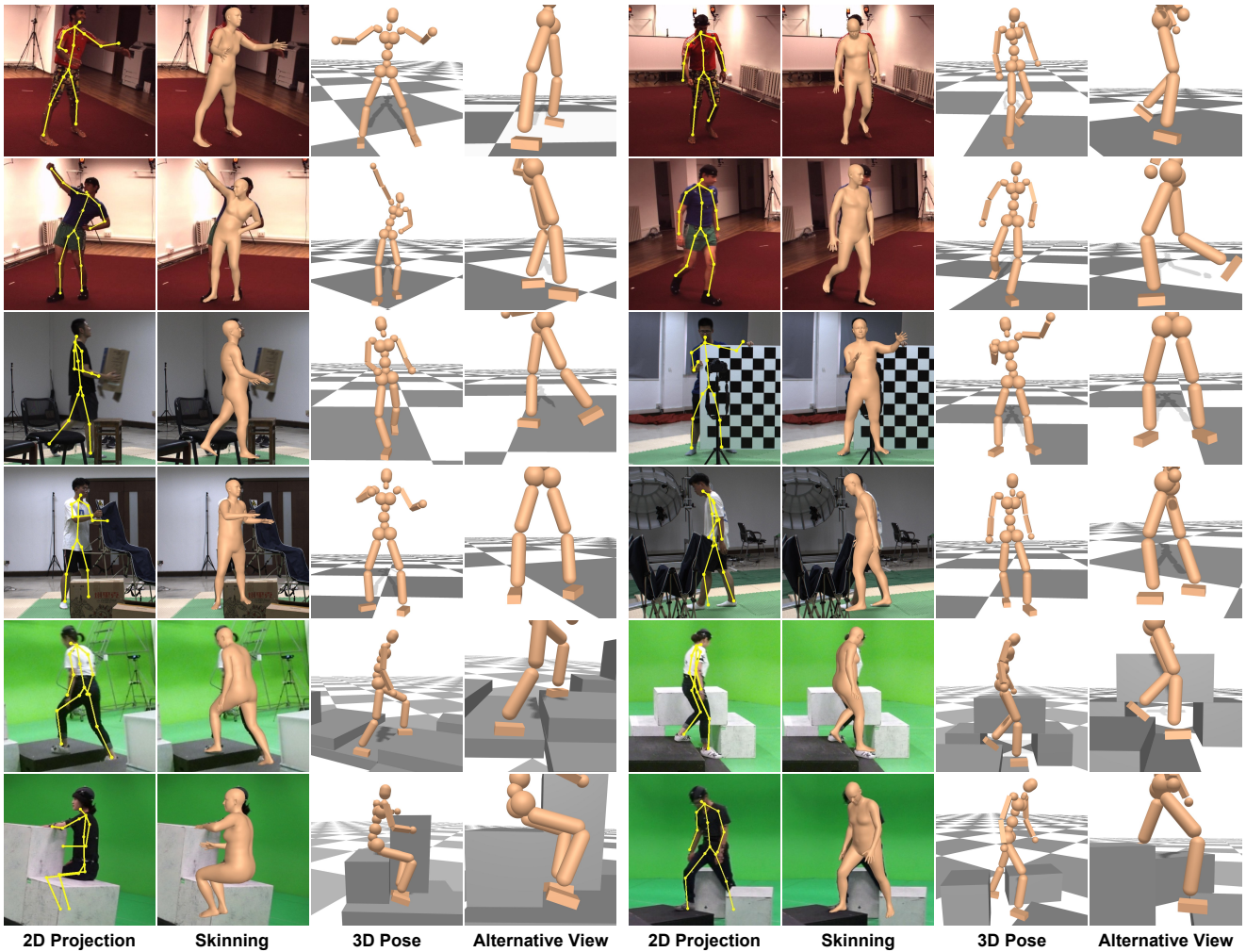}
    \end{center}
    \vspace{-5mm}
    \caption{Qualitative results on Human3.6M~(row 1-2), 3DOH~(row 3-4) and GPA~(row 5-6) dataset. Our method is robust to complex terrains, occlusions and body shape variations. We can obtain natural skinning mesh with the estimated pose parameters.}
    \label{fig:results}
    \vspace{-5mm}
    \end{figure*}

\begin{table}
    \begin{center}
        \resizebox{1.0\linewidth}{!}{
            \begin{tabular}{l c c c c c c}
            \noalign{\hrule height 1.5pt}
            \multirow{2}{*}{Joint} &\multirow{2}{*}{-x} &\multirow{2}{*}{+x} &\multirow{2}{*}{-y} &\multirow{2}{*}{+y} &\multirow{2}{*}{-z} &\multirow{2}{*}{+z}\\
            &\\
            \noalign{\hrule height 1pt}

            Left Hip   &-2.0 &2.0 &-0.57 &0.57 &-0.27 &0.27\\
            Left Knee  &-0.3 &1.57 &-0.27 &0.27 &-0.0 &0.0\\
            Left Ankle  &-0.57 &0.57 &-0.57 &1.2 &-0.57 &0.57\\
            Right Hip    &-2.0 &2.0 &-0.57 &0.57 &-0.27 &0.27\\
            Right Knee   &-0.3 &1.57 &-0.27 &0.27 &-0.0 &0.0\\
            Right Ankle  &-0.57 &0.57 &-1.2 &0.57 &-0.57 &0.57\\

            Lower Back     &-1.57 &1.57 &-1.57 &1.57 &-1.57 &1.57\\
            Upper Back     &-1.57 &1.57 &-1.57 &1.57 &-1.57 &1.57\\
            Chest          &-1.57 &1.57 &-1.57 &1.57 &-1.57 &1.57\\
            Lower Neck     &-0.57 &0.57 &0.0 &0.0 &0.0 &0.0\\
            Upper Neck     &-0.57 &0.57 &-0.57 &0.57 &0.0 &0.0\\

            Left Clavicle  &-1.57 &1.57 &-1.57 &1.57 &-1.57 &1.57\\
            Left Shoulder  &-1.57 &1.57 &-1.57 &1.57 &-1.57 &1.57\\
            Left Elbow     &-1.57 &1.57 &-1.57 &1.57 &-1.57 &1.57\\
            Right Clavicle &-1.57 &1.57 &-1.57 &1.57 &-1.57 &1.57\\
            Right Shoulder &-1.57 &1.57 &-1.57 &1.57 &-1.57 &1.57\\
            Right Elbow    &-1.57 &1.57 &-1.57 &1.57 &-1.57 &1.57\\
            \noalign{\hrule height 1.5pt}
            \end{tabular}
        }
    \end{center}
    \vspace{-6mm}
    \caption{The limitations of joint rotations for CMA-ES. The unit of the numbers is radian.}
    \label{tab:bounds}
    \vspace{-2mm}
\end{table}

\begin{table}
    \begin{center}
        \resizebox{1.0\linewidth}{!}{
            \begin{tabular}{l c c c c c c}
            \noalign{\hrule height 1.5pt}
            \multirow{2}{*}{Stage} &\multirow{2}{*}{data} &\multirow{2}{*}{latent prior} &\multirow{2}{*}{shape prior} &\multirow{2}{*}{kinetic prior} &\multirow{2}{*}{interaction term} \\
            &\\
            \noalign{\hrule height 1pt}

            Stage1   &1.0 &4040.0 &100.0 &1000.0 &0.0 \\
            Stage2  &1.0 &404.0 &50.0 &500.0 &0.0 \\
            Stage3  &1.0 &57.4 &10.0 &250.0 &0.0 \\
            Stage4    &1.0 &1.78 &5.0 &200.0 &4500.0 \\
            \noalign{\hrule height 1.5pt}
            \end{tabular}
        }
    \end{center}
    \vspace{-5mm}
    \caption{The loss weights of kinematic optimization in each stage.}
    \label{tab:optimization}
    \vspace{-5mm}
\end{table}

\subsection{Training details}\label{sec:trainingdetails}
We introduce the training details of our distribution prior in this section. As mentioned in the main paper, the train set from Human3.6M and GPA are used for training. We first apply the CMA-ES method to get pseudo ground-truth. Since generating sampled target pose for a complete motion sequence is difficult and time-consuming, we select two consecutive frames from the dataset and calculate the state of character from the dataset with linear interpolation. The kinematic pose in the second frame is used as reference. We then apply CMA-ES method to obtain the target pose distribution. When the prior is convergent, we finish the pre-train procedure and incorporate the two-branch decoder to refine the network. It is an ideal situation to have the character state as the same as the state from linear interpolation. In the training phase, we add random noises in the character state to simulate the discrepancy of real simulation. The distribution prior is trained on a single NVIDIA TITAN RTX GPU with a learning rate of 0.0001 and a batch size of 32.

\subsection{Sampling details}\label{sec:samplingdetails}
We describe implementation details of the neural motion control. The character state $\boldsymbol{s}$ consists of pose $\boldsymbol{q}$ and velocity $\dot{\boldsymbol{q}}$. The first 6 dimensions of pose are global translation and global rotation. The rest are joint rotations that are represented by axis-angle. Besides, $\dot{\boldsymbol{q}}$ contains 3-dimension base linear velocity, 3-dimension base angular velocity and 51-dimension joint angular velocity. The total dimension of character state is 114. In the physics simulator, the rotations are represented by quaternion. 

We do not directly control the root joint, thus the target pose has the same dimension as the DOF~(degree-of-freedom) of moveable joints, which is 51. Given a target pose, we first compute torques from the PD controller and limit the torques in a reasonable range. The parameters are shown in~\cref{tab:control}. Finally, the torques are applied to the character via torque control mode. We simulate 8 times for a given target pose and re-calculate the torques based on the target pose and simulated pose in each time. When applying the distribution prior in neural motion control, we use the first frame of reference motion to initialize the physical character.

\subsection{Optimization details}\label{sec:optimizationdetails}
Our optimization has four stages. The only difference among each stage is the loss weights for each term. The different loss weights promote the optimized results from coarse to fine. As shown in~\cref{tab:optimization}, the optimization with the weights in the first three stages can obtain proximate results. We only apply the interaction constraint in the last stage to get accurate ground contact.

\subsection{Details on SDF}\label{sec:sdf}
The SDF representation is similar to \cite{hassan2019resolving}, in which the scene is used to constrain a single human pose. We use a uniform voxel grid with the size 256$\times$256$\times$256 to represent the field. The trilinear interpolation is used for the discretization of the 3D distance field with the limited grid resolution. The resolution is enough to obtain coarse contact for physics-based motion capture with reasonable computational complexity and memory consumption.

\section{Success rate}\label{sec:Success}
Since our method is based on sampling, the reconstructed control is not guaranteed to be successful after a single run of the sampling algorithm~\cite{liu2010sampling}. The character may fall and be unable to finish the complete motion. Thus, the success rate is an important metric to evaluate our method. For sampling-based motion control, the sampling distribution is the most important influencing factor for the metric of success rate. CMA-ES-based method~\cite{liu2015improving} learn from previous trials to update the distribution via online adaptation and might require more trials for motion capture tasks. The initial several trials draw samples randomly and blindly and are very likely to fail, which results in a low overall success rate. The limitation is also demonstrated in a recent work~\cite{xie2021inverse}. With the well-trained prior, our method samples from the regressed distribution and has a high success rate for all trials. In~\cref{tab:success}, we follow~\cite{xie2021inverse} and~\cite{liu2010sampling} to conduct a comparison on the lift leg motion. The success rate is 97\% and 90\% for our method and~\cite{liu2015improving}, but \cite{liu2010sampling} is 83\%. The comparisons illustrate the advantage of our approach from a different perspective. In addition, to increase the number of samples and saved samples at each iteration can improve the success rate. Furthermore, when the tracking is fail, we can also run multiple times on the same problem to allow a user to explore different possible reconstructions.

\begin{table}
    \begin{center}
        \resizebox{1.0\linewidth}{!}{
            \begin{tabular}{l c c c c c c}
            \noalign{\hrule height 1.5pt}
            Method &Liu~\etal~\cite{liu2010sampling} &Liu~\etal~\cite{liu2015improving} &Neural MoCon \\
            \noalign{\hrule height 1pt}
            Success rate   &83\% &90\% &97\% \\
            \noalign{\hrule height 1.5pt}
            \end{tabular}
        }
    \end{center}
    \vspace{-5mm}
    \caption{The comparison on the lift leg motion with the metric of success rate.}
    \label{tab:success}
    \vspace{-5mm}
\end{table}

\section{More results and discussions}\label{sec:experiments}

We show more qualitative results in this section to demonstrate the performance of our method. In~\cref{fig:results}, the results on GPA, 3DOH and Human3.6M dataset show that our method is robust to complex terrains, occlusions and body shape variations. Furthermore, we apply the estimated pose from neural motion control to SMPL model and get the skinning mesh. It shows the obtained meshes are natural and accurate. Since the collision detection is conducted on the primitives of physical character, the shape discrepancies of hand and foot cause a slight interpenetration on the skinning mesh. We will detect mesh-level collision or design more delicate characters to prevent these artifacts in the future work.  

\section{Video}\label{sec:video}
In the video, we show the qualitative comparisons with VIBE, DMMR, and PhysCap. Due to the hard physical constraints, our method can prevent floor interpenetration. Most of the foot sliding is also avoided by applying lateral friction. To demonstrate the performance of our method on complex terrain, we conducted a comparison with PhysCap on the GPA dataset. We used the original character model of PhysCap. Although PhysCap can obtain smooth motion, the wrong contact states for uneven terrain scenario result in floating motion. However, since the interaction constraint is used in kinematic optimization, our method can produce a physically plausible and high-quality motion with the proposed neural motion control.

\section{Why neural motion control?}\label{sec:why}
To build a 3D human dataset with accurate force annotations is complex and expensive~\cite{zell2020weakly}. The joint torques cannot be measured non-intrusively and therefore need to be derived using computationally expensive optimization techniques. Furthermore, the torques for different subjects with different body shapes have large variances. It results in a poor generalization ability for the network that directly regresses joint torques. However, the distribution prior is a network that regresses the target pose distribution. With the dense supervision from pseudo ground truth and the two-branch decoder, the network is easy to be convergent. In addition, with the sampling, the neural motion control is more general to complex terrain, body shape variations, and diverse behaviors.

Compared to CMA-ES based method, the existing sampling-based motion control first relies on CMA-ES to adapt the distribution via evaluating plenty of samples, which is time-consuming. Our network-based prior avoids such distribution adaptation, and elite samples can be directly obtained from the regressed distribution. Our method saves a lot of sample evaluations compared to~\cite{liu2015improving}. Furthermore, the distribution adaptation relies on random samples from an initial distribution to update the distribution via CMA-ES, which imposes uncertainty for the motion capture. The proposed prior avoids the uncertainty, and the precise control can be acquired by sampling from the distribution of network output, which is the same as CMA-ES-based approaches. Combing neural networks and sampling-based motion control provides a feasible solution to achieve real-time physics-based motion capture though there is still a gap for this goal.

\end{document}